\documentclass[11pt]{article}

\usepackage[T1]{fontenc}
\usepackage[letterpaper,margin=1in]{geometry}
\usepackage{amsmath,amssymb,mathtools,bm}
\usepackage{graphicx}
\usepackage{booktabs,multirow,array,tabularx}
\usepackage{placeins}
\usepackage{xcolor}
\usepackage{tikz}
\usepackage{url}
\usepackage[hidelinks]{hyperref}
\usetikzlibrary{arrows.meta,positioning,fit,backgrounds,shapes.geometric,decorations.pathreplacing,angles,quotes,calc}

\newcommand{\R}{\mathbb{R}}
\newcommand{\FCP}{\operatorname{FCP}}

\newcommand{\Span}{\operatorname{span}}

\newcommand{\eps}{\varepsilon}
\newcommand{\norm}[1]{\left\lVert#1\right\rVert}
\newcommand{\abs}[1]{\left\lvert#1\right\rvert}

\begin{document}

\title{Fractal Cross Product: Theory, Differentiable Implementation,\\
and Application to Medical Image Analysis}

\author{Noaman Khan \quad Nihad hadj sahraoui \quad Samir Brahim Belhaouari \\[0.5em]
\normalsize College of Science and Engineering, Hamad Bin Khalifa University, Doha, Qatar\\
\normalsize \texttt{nokh88609@hbku.edu.qa; nihadhsa@gmail.com; sbelhaouari@hbku.edu.qa}}
\date{}
\maketitle

\begin{abstract}
The magnitude of the generalized Euclidean cross product is a Gram volume
whose degree under common scaling is fixed by the integer dimension of the
spanning frame. We formulate a generalized Fractal Cross Product (FCP) as a
nonlinear radial deformation with a prescribed positive degree $D$, which may
be non-integer. The scalar construction applies in any ambient dimension
$m\geq k$, while its canonically oriented vector form requires codimension
one. It recovers the classical generalized cross product exactly at $D=k$ and
retains orthogonality, alternation, rotation equivariance, and
$D$-homogeneity, but is generally not multilinear. For exact self-similar
frame systems, the construction also obeys a scale-balance law at the
similarity dimension. We derive a differentiable, dimensionless image
response and a non-circular empirical accumulation exponent obtained by
regressing raw angular Gram responses across patch widths. Binary64
calculations recover the finite-frame identities to roundoff, while raster
experiments recover the Sierpi\'nski-triangle value 1.5849625 at three
resolutions. In five-seed medical-imaging comparisons, FCP-centered fusion
increased mean area under the receiver operating characteristic curve from
0.7264 to 0.8135 and from 0.5843 to 0.6765 on the random and
hospital-separated Retinal Image Database for Optic Nerve Evaluation
partitions, respectively, and from 0.7403 to 0.7475 on FracAtlas. On
FracAtlas, balanced accuracy increased from 0.6186 to 0.6721 and the harmonic
mean of precision and sensitivity from 0.3669 to 0.4457. These results support
the utility of the complete fusion framework, but do not isolate the effect of
FCP from that of its complementary descriptors and fusion head.
\end{abstract}


\section{Introduction}

For two vectors in $\R^3$, the magnitude of the classical cross product equals
the area of the parallelotope they span. More generally, the norm of the Hodge
dual of a $k$-fold exterior product is the $k$-dimensional Gram volume of the
frame \cite{Flanders1989,HornJohnson2013,Belhaouari2025}. Common dilation by a
factor $\lambda$ therefore scales this volume by $\abs{\lambda}^k$: its
homogeneity is fixed by the integer frame dimension.

This fixed degree motivates the mathematical question considered here: can a
Gram-volume construction be deformed so that its magnitude has an arbitrary
prescribed degree $D>0$, while retaining the orientation and orthogonality of
the generalized cross product in codimension one? The question is relevant to
self-similar geometry, where a system of $N$ copies with contraction ratio $r$
has similarity dimension $D$ satisfying $Nr^D=1$ under the usual separation
conditions \cite{Hutchinson1981,Falconer2014}. It does not follow, however,
that introducing a non-integer exponent creates a fractal measure. Hausdorff
measures and related calculi require limiting constructions
\cite{Federer1969,Mattila1995,Parvate2011,WinterZahle2013}.

We address this question through a power of the Gram determinant. Within this
specific ansatz, the homogeneity requirement determines the exponent. The
resulting scalar and vector forms are
\begin{equation}
M_D(\bm v_1,\ldots,\bm v_k)=(\det G)^{D/(2k)},
\label{eq:intro_md}
\end{equation}
and, when the ambient dimension is $k+1$, the oriented vector construction
\begin{equation}
\FCP_D(\bm v_1,\ldots,\bm v_k)
=\norm{X}^{D/k-1}X,
\quad X=\star(\bm v_1\wedge\cdots\wedge\bm v_k).
\label{eq:intro_fcp}
\end{equation}
The zero-frame case is defined separately. The scalar form applies for
$m\geq k$, whereas the canonical vector form requires $m=k+1$. FCP is thus a
nonlinear, $D$-homogeneous radial deformation that recovers the classical
generalized cross product at $D=k$ and loses multilinearity otherwise. The
term \emph{fractal} refers specifically to the balance obtained for exact
self-similar frame systems at $Nr^D=1$; it does not identify the finite-frame
functional with Hausdorff measure or make it a general estimator of fractal
dimension.

Fractional coordinate products and vector calculus in non-integer-dimensional
spaces address related but distinct questions \cite{KankarejSingh2023,
Tarasov2015}. To the best of our knowledge, existing work does not combine the
same Gram-volume power parameterization, prescribed homogeneity degree,
codimension-one Hodge orientation, and exact recovery at $D=k$. This claim
concerns the combined formulation, not the novelty of fractional powers, Gram
determinants, or exterior algebra.

For image analysis, the scalar construction avoids choosing a nonunique normal
in high-dimensional feature space. Its normalized form becomes a
dimensionless angular response rather than the exact length-homogeneous FCP.
We also derive an empirical accumulation exponent from explicit log-log
regression of raw multiscale responses, thereby avoiding insertion of a target
$D$ before estimating scale behavior. The practical evaluation uses canonical
binary sets, including standard fractals, and two medical-image classification
tasks. Fractal descriptors have been studied extensively in retinal fundus
analysis
\cite{Family1989,Masters2004,Stosic2006,LopesBetrouni2009}.

The contributions are:
\begin{enumerate}
\item the scalar Gram functional and its codimension-one oriented FCP, with a
derivation and precise classical limit;
\item an analysis of the inherited geometric properties,
non-multilinearity, conditional self-similar balance, and numerical stability;
\item a stable normalized response, a non-circular image-level FCP
accumulation exponent obtained from explicit log-log scale regression, and
differentiable dense and scale-regression realizations; and
\item validation on canonical sets and two five-seed medical-imaging
applications in retinal fundus analysis and fracture radiography, evaluating
FCP-centered fusion with CNNs trained from scratch.
\end{enumerate}

\section{Mathematical Background and Related Work}
\label{sec:background}

\subsection{Notation}

Throughout, $k$ is the number of spanning vectors, $m$ is the ambient
Euclidean dimension, and $D>0$ is a prescribed scaling exponent. Thus
\begin{equation}
\bm v_1,\ldots,\bm v_k\in\R^m,\qquad m\ge k.
\end{equation}
The matrix $V\in\R^{m\times k}$ has columns $\bm v_i$, and
$G=V^\top V\in\R^{k\times k}$ is its Gram matrix. The vector-valued
construction requires $m=k+1$; the scalar construction does not. Bold symbols
denote vectors and matrices, $\norm{\cdot}$ is the Euclidean norm, $O(m)$ and
$SO(m)$ are the orthogonal and proper-orthogonal groups, $\wedge$ is the
exterior product, and $\star$ is the Hodge star associated with the standard
orientation.

\subsection{Euclidean Geometry, Gram Volume, and Generalized Cross Products}

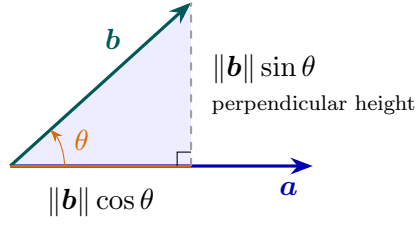
\begin{figure}[t]
\centering
\begin{tikzpicture}[x=0.92cm,y=0.92cm,>=Stealth]
  \coordinate (O) at (0,0);
  \coordinate (A) at (4.35,0);
  \coordinate (H) at (2.60,0);
  \coordinate (B) at (2.60,2.35);
  \fill[blue!7] (O)--(H)--(B)--cycle;
  \draw[->,very thick,blue!70!black] (O)--(A)
    node[pos=0.92,below=2pt] {$\bm a$};
  \draw[->,very thick,teal!70!black] (O)--(B)
    node[pos=0.66,above left=-1pt] {$\bm b$};
  \draw[dashed,thick,gray!75] (H)--(B);
  \draw[thick,orange!85!black] (O)--(H);
  \draw (H) ++(-0.20,0) -- ++(0,0.20) -- ++(0.20,0);
  \draw[->,orange!85!black] (0.78,0) arc[start angle=0,end angle=42.1,radius=0.78];
  \node[orange!85!black] at (1.03,0.37) {$\theta$};
  \node[below=4pt] at ($(O)!0.5!(H)$) {$\norm{\bm b}\cos\theta$};
  \node[right=4pt,align=left] at ($(H)!0.5!(B)$)
    {$\norm{\bm b}\sin\theta$\\[-1pt]\scriptsize perpendicular height};
  \node[above=2pt] at ($(O)!0.56!(B)$) {};
\end{tikzpicture}
\caption{Triangle geometry underlying the two-vector construction. The
projection of $\bm b$ onto $\bm a$ has length $\norm{\bm b}\cos\theta$ and
the perpendicular component has length $\norm{\bm b}\abs{\sin\theta}$, so
$\norm{\bm a\times\bm b}$ is base times height.}
\label{fig:triangle_geometry}
\end{figure}

For $\bm a,\bm b\in\R^3$ meeting at angle $\theta$, the familiar relations
\begin{equation}
\bm a\cdot\bm b=\norm{\bm a}\norm{\bm b}\cos\theta,
\qquad
\norm{\bm a\times\bm b}
=\norm{\bm a}\norm{\bm b}\abs{\sin\theta}
\label{eq:dot_cross}
\end{equation}
show that the cross-product magnitude is the product of a base length and its
perpendicular height. The corresponding triangle has one half of this area
(Fig.~\ref{fig:triangle_geometry}).

For a general $k$-frame, the analogous volume is encoded by the Gram matrix.
Given $\bm v_1,\ldots,\bm v_k\in\R^m$, define
\begin{equation}
G=\begin{bmatrix}
\bm v_1\cdot\bm v_1&\cdots&\bm v_1\cdot\bm v_k\\
\vdots&\ddots&\vdots\\
\bm v_k\cdot\bm v_1&\cdots&\bm v_k\cdot\bm v_k
\end{bmatrix}=V^\top V.
\label{eq:gram}
\end{equation}
The singular-value decomposition $V=U\Sigma W^\top$ gives
$G=W\Sigma^2W^\top$ and hence
\begin{equation}
\boxed{V_k(\bm v_1,\ldots,\bm v_k)=\sqrt{\det G}
=\prod_{i=1}^{k}\sigma_i(V).}
\label{eq:gram_volume}
\end{equation}
This is the Euclidean $k$-volume of the spanned parallelotope
\cite{HornJohnson2013}. When $m=k+1$, exterior algebra supplies an oriented
normal
\begin{equation}
\boxed{X(\bm v_1,\ldots,\bm v_k)
=\star(\bm v_1\wedge\cdots\wedge\bm v_k),}
\label{eq:general_cross}
\end{equation}
with $\norm{X}=\sqrt{\det G}$ and
$X\perp\Span\{\bm v_1,\ldots,\bm v_k\}$
\cite{Flanders1989,Belhaouari2025}. Codimension one is essential: if
$k<m-1$, the orthogonal complement has dimension greater than one and no
unique normal direction follows from the frame alone.

\subsection{Self-Similar Scaling and Fractal Measures}

For an iterated function system of $N$ similitudes with common contraction
$r\in(0,1)$, the similarity dimension is
\begin{equation}
Nr^D=1,
\qquad
D=\frac{\log N}{\log(1/r)}.
\label{eq:similarity_dimension}
\end{equation}
Under the open-set condition this dimension agrees with the Hausdorff
dimension of the invariant set \cite{Hutchinson1981,Falconer2014}. Outside
this setting, box, Hausdorff, and similarity dimensions need not agree and are
therefore kept distinct in this paper \cite{Mattila1995}. Multifractal
analysis replaces a single exponent with a spectrum of local singularities
\cite{ChhabraJensen1989}; FCP instead uses one prescribed exponent per
evaluation and does not constitute a multifractal formalism.

Hausdorff measure is constructed through infima over covers whose diameters
tend to zero; Minkowski content is defined through asymptotic volumes of
parallel neighborhoods \cite{Federer1969,Mattila1995}. Fractal curvature and
curvature-direction measures refine such asymptotics for suitable
self-similar sets \cite{WinterZahle2013,BohlZahle2013}. Other programs define
calculus along fractal curves or use a non-integer-dimensional ambient-space
formalism \cite{Parvate2011,Tarasov2015}. These constructions involve limits,
measures, or differential operators. By contrast, $M_D$ is defined below as a
finite-frame functional. Its exponent is motivated by fractal scaling, but
identification with an established fractal measure would require a separate
convergence analysis.

\subsection{Related Fractional Vector Formulations}

Kankarej and Singh's fractional vector product is a bilinear coordinate
interpolation in $\R^3$ whose unit-parameter case recovers the ordinary cross
product \cite{KankarejSingh2023}. The present proposal solves a different
problem: its parameter is a homogeneity degree, its magnitude is a power of a
Gram determinant, and it becomes non-multilinear except at the Euclidean
degree.

\subsection{Fractal Descriptors and Medical Imaging}

Fractal and multifractal descriptors have been used to quantify irregularity,
branching, and texture across medical-imaging modalities
\cite{LopesBetrouni2009}. Retinal vasculature is a prominent example:
statistical self-similarity and fractal dimension have been studied in normal
retinae and in vascular disease \cite{Family1989,Masters2004,Stosic2006,
Grauslund2010}. This literature motivates evaluating a scale-sensitive
geometric representation on retinal images, without implying that a learned
neural exponent is a physical vascular dimension. Medical-image models may
also exploit acquisition- or site-specific effects; hospital-separated
partitions and leakage-separated model selection are therefore relevant to
the evaluation \cite{Zech2018,Yu2022,Kelly2019}.

\section{Generalized Fractal Cross Product: Mathematical Analysis}
\label{sec:theory}

\subsection{Problem Formulation and Derivation}

Let $k$ vectors span a Euclidean parallelotope. We seek a Gram-based functional
whose magnitude has prescribed common-dilation degree $D>0$ and whose oriented
codimension-one realization agrees with the generalized Euclidean cross
product at the Euclidean degree. The definition should remain meaningful when
$D$ is non-integer. The Euclidean Gram volume obeys
\begin{equation}
V_k(\lambda\bm v_1,\ldots,\lambda\bm v_k)
=\abs{\lambda}^{k}V_k(\bm v_1,\ldots,\bm v_k).
\label{eq:euclidean_homogeneity}
\end{equation}
Restricting attention to the Gram-power ansatz
$M_D=(\det G)^\alpha$, common scaling sends $G\mapsto\lambda^2G$ and therefore
\begin{equation}
\det G\mapsto\abs{\lambda}^{2k}\det G,
\qquad
(\det G)^\alpha\mapsto
\abs{\lambda}^{2k\alpha}(\det G)^\alpha.
\end{equation}
Matching the desired degree requires $2k\alpha=D$, so within this ansatz
\begin{equation}
\boxed{\alpha=\frac{D}{2k}.}
\label{eq:alpha_derivation}
\end{equation}

\subsection{FCP Definition and Relation to the Classical Product}

For $\bm v_1,\ldots,\bm v_k\in\R^m$, $m\geq k$, and $D>0$, define the scalar
FCP functional
\begin{equation}
\boxed{M_D(\bm v_1,\ldots,\bm v_k)
=(\det G)^{D/(2k)}=V_k^{D/k}.}
\label{eq:md}
\end{equation}
Positive semidefiniteness of $G$ makes $M_D$ nonnegative, and $M_D$ vanishes
exactly when the frame is linearly dependent. Orthogonal changes of coordinates
preserve $G$, whereas common dilation gives
\begin{equation}
\boxed{M_D(\lambda\bm v_1,\ldots,\lambda\bm v_k)
=\abs{\lambda}^{D}M_D(\bm v_1,\ldots,\bm v_k).}
\label{eq:md_homogeneity}
\end{equation}
These properties follow directly from
$\det(\lambda^2G)=\abs{\lambda}^{2k}\det G$. Equation~\eqref{eq:md} defines a
finite-frame functional rather than a countably additive measure.

A canonical oriented vector is available only in codimension one. For
$\bm v_1,\ldots,\bm v_k\in\R^{k+1}$, let
$X=\star(\bm v_1\wedge\cdots\wedge\bm v_k)$ and define
\begin{equation}
\boxed{
\FCP_D(\bm v_1,\ldots,\bm v_k)=
\begin{cases}
\norm{X}^{D/k-1}X,&X\ne\bm0,\\[2pt]
\bm0,&X=\bm0.
\end{cases}}
\label{eq:fcp_definition}
\end{equation}
Thus $\norm{\FCP_D}=M_D$ and, for a nondegenerate frame,
$\FCP_D=M_DX/\norm{X}$. The Hodge-star direction is orthogonal to the input
span and reverses sign under an odd permutation. FCP inherits these geometric
properties from $X$; the deformation changes its magnitude but not its
orientation.

The relationship with the Euclidean generalized cross product is exact only
when the prescribed degree equals the frame dimension:
\begin{equation}
\boxed{\FCP_k(\bm v_1,\ldots,\bm v_k)
=\star(\bm v_1\wedge\cdots\wedge\bm v_k)=X.}
\label{eq:classical_limit}
\end{equation}
Under a common real dilation, the magnitude satisfies
\begin{equation}
\boxed{\begin{aligned}
&\norm{\FCP_D(\lambda\bm v_1,\ldots,\lambda\bm v_k)}\\
&\quad=\abs{\lambda}^{D}\norm{\FCP_D(\bm v_1,\ldots,\bm v_k)}.
\end{aligned}}
\label{eq:magnitude_scaling}
\end{equation}
For $\lambda>0$, the vector itself gains the factor $\lambda^D$; for
$\lambda<0$, the oriented factor is $(-1)^k\abs{\lambda}^D$. For
$Q\in SO(k+1)$,
\begin{equation}
\FCP_D(Q\bm v_1,\ldots,Q\bm v_k)
=Q\FCP_D(\bm v_1,\ldots,\bm v_k),
\label{eq:rotation_equivariance}
\end{equation}
while scaling one argument by $c>0$ gives a factor $c^{D/k}$. The construction
is therefore rotation equivariant but cannot be multilinear unless $D=k$.
Accordingly, FCP is a nonlinear $D$-homogeneous deformation of the generalized
Euclidean cross product, not a new multilinear product.

\subsection{Two-Vector Triangle Formula}

For $\bm a,\bm b\in\R^3$,
\begin{equation}
G=\begin{bmatrix}
\norm{\bm a}^2&\bm a\cdot\bm b\\
\bm a\cdot\bm b&\norm{\bm b}^2
\end{bmatrix}.
\end{equation}
Using \eqref{eq:dot_cross},
\begin{align}
\det G
&=\norm{\bm a}^2\norm{\bm b}^2-(\bm a\cdot\bm b)^2\\
&=\norm{\bm a}^2\norm{\bm b}^2(1-\cos^2\theta)\\
&=\boxed{\norm{\bm a}^2\norm{\bm b}^2\sin^2\theta}.
\label{eq:triangle_det}
\end{align}
Therefore
\begin{equation}
\boxed{\norm{\FCP_D(\bm a,\bm b)}
=\bigl(\norm{\bm a}\norm{\bm b}\abs{\sin\theta}\bigr)^{D/2}.}
\label{eq:triangle_fcp}
\end{equation}
At $D=2$, this equals $\norm{\bm a\times\bm b}$. If
$\norm{\bm a}=\norm{\bm b}=s$, then
\begin{equation}
\norm{\FCP_D(\bm a,\bm b)}
=s^D\abs{\sin\theta}^{D/2},
\label{eq:equal_side}
\end{equation}
whereas the ordinary cross-product magnitude is
$s^2\abs{\sin\theta}$. The deformation changes both length scaling and the
angular exponent; replacing \eqref{eq:equal_side} by
$s^D\sin\theta$ would define a different operator.

\subsection{Conditional Self-Similar Scaling}

The link with fractal scaling requires an explicitly defined family of local
frames; it does not follow from a non-integer value of $D$ alone.

\paragraph*{Conditional self-similar scaling observation.}
Let an exact self-similar construction have $N$ congruent copies with common
contraction ratio $r\in(0,1)$. Assume that every copy contains a corresponding
local $k$-frame obtained from a reference frame by scaling by $r$ and applying
an orthogonal map; translations are irrelevant. If $D>0$ satisfies $Nr^D=1$,
then the sum of the FCP magnitudes of the corresponding frames is invariant
between successive construction levels. At level $q$,
\begin{equation}
\boxed{F_q=N^q r^{qD}F_0=(Nr^D)^qF_0=F_0.}
\label{eq:selfsimilar_conservation}
\end{equation}
By \eqref{eq:magnitude_scaling} and orthogonal invariance of the magnitude, one
copy contributes $r^DF_0$. Summing the $N$ congruent copies gives
$F_1=Nr^DF_0=F_0$. At level $q$ there are $N^q$ corresponding frames, each
scaled by $r^q$ and hence contributing $r^{qD}F_0$. Equation
\eqref{eq:selfsimilar_conservation} follows.

The assumptions are substantive. Frames must correspond across exact
similarity copies, and their magnitudes must be accumulated using the stated
normalization. The observation therefore does not extend automatically to
random, overlapping, anisotropic, or multifractal constructions.

\subsection{Sierpi\'nski Triangle and the Wrong-Dimension Law}

For the Sierpi\'nski triangle, $N=3$, $r=1/2$, and
\begin{equation}
D_S=\frac{\log3}{\log2}\approx1.5849625.
\label{eq:sierpinski_dimension}
\end{equation}
Each level-one frame contributes $2^{-D_S}F_0$, so
$F_1=3\,2^{-D_S}F_0=F_0$; at level $q$,
$F_q=3^q2^{-qD_S}F_0=F_0$.

For a trial exponent $D'$, the same argument yields
\begin{equation}
\boxed{F_q(D')=(Nr^{D'})^qF_0(D').}
\label{eq:wrong_d}
\end{equation}
Therefore $F_q(D')\to0$ if $Nr^{D'}<1$, diverges if $Nr^{D'}>1$, and remains
constant at $Nr^{D'}=1$. Because $r<1$, Sierpi\'nski trials below $D_S$ grow
and trials above $D_S$ decay. The critical crossing could motivate an
FCP-based similarity-dimension estimator, but no finite-sample consistency or
robustness result is claimed.

\subsection{Relationship to Hausdorff Measure}

The homogeneity law in \eqref{eq:md_homogeneity} has the dilation degree
expected of a $D$-dimensional measure, but homogeneity alone is insufficient
for a measure-theoretic identification. Proving convergence
to or equivalence with Hausdorff measure would require: (i) a specified
sequence of geometric approximations or covers; (ii) a local-frame selection
rule; (iii) normalization preventing overcounting; (iv) a limit as scale tends
to zero; (v) existence of the limiting measure; and, for equivalence, (vi)
comparison constants \cite{Federer1969,Mattila1995}. The present construction
provides a frame functional and a conditional self-similar balance identity;
it does not establish this measure-theoretic program.

\subsection{Scope and Numerical Stability}
\label{sec:stability}

Directly evaluating $\det G$ can lose relative accuracy for nearly dependent
frames. If $\sigma_i$ are the singular values of $V$, then
\begin{equation}
M_D=\left(\prod_{i=1}^k\sigma_i\right)^{D/k},
\qquad
\log M_D=\frac{D}{k}\sum_{i=1}^{k}\log\sigma_i.
\label{eq:svd_stable}
\end{equation}
For ill-conditioned frames, an SVD is generally preferable to direct
determinant evaluation; Cholesky or log-determinant methods are suitable when
$G$ is positive definite \cite{Higham2002}. Degenerate frames are assigned
zero, and numerical orientation must be fixed by signed minors or an oriented
factorization. These conditions delimit the geometric construction used in
the image formulation below: its scalar form extends to arbitrary ambient
dimension, but its canonical oriented vector form does not.

\section{Differentiable FCP for Image Analysis}
\label{sec:neural}

The exact geometric construction acts on finite vector frames, whereas image
features require a scalar realization in a potentially high-dimensional
ambient space. To keep these objects distinct, we use the following three
levels throughout the remainder of the paper.
\begin{table}[t]
\centering
\caption{Three levels of the FCP construction.}
\label{tab:levels}
\small
\begin{tabularx}{\columnwidth}{@{}p{0.16\columnwidth}X@{}}
\toprule
\textbf{Level} & \textbf{Object and scope}\\
\midrule
I & Exact vector $\FCP_D$ in codimension one; oriented and
$D$-homogeneous.\\
II & Scalar $M_D=(\det G)^{D/(2k)}$ for arbitrary $m\ge k$; orthogonally
invariant and $D$-homogeneous.\\
III & Normalized neural response $R_D=\abs{\sin\theta}^{D/2}$ for two
nonzero feature vectors; dimensionless and not length-homogeneous.\\
\bottomrule
\end{tabularx}
\end{table}
Properties of the exact FCP are not attributed to the normalized response
unless they remain valid after normalization.

\subsection{Scalar Gram Construction in Arbitrary Feature Dimension}

Suppose a local image patch produces feature vectors
$\bm u,\bm v\in\R^w$. The number of vectors does not determine the ambient
dimension. When $w>3$ and the vectors are linearly independent, their
orthogonal complement has dimension $w-2$; it is larger for a dependent pair.
No unique normal vector is therefore available, but the Level-II scalar
construction remains well defined:
\begin{equation}
\boxed{M_D(\bm u,\bm v)=
\left[\norm{\bm u}^2\norm{\bm v}^2-(\bm u^\top\bm v)^2\right]^{D/4}.}
\label{eq:image_scalar}
\end{equation}
The image implementation consequently uses this scalar quantity rather than
selecting an arbitrary normal from a $(w-2)$-dimensional normal space.

\subsection{Normalized FCP Response}

To match the implementation exactly, define
\begin{equation}
\widehat{\bm u}=\frac{\bm u}{\max\{\norm{\bm u}_2,\eps\}},
\qquad
\widehat{\bm v}=\frac{\bm v}{\max\{\norm{\bm v}_2,\eps\}},
\end{equation}
and
\begin{equation}
c=\operatorname{clip}(\widehat{\bm u}^{\top}\widehat{\bm v},-1,1),
\end{equation}
and let $\chi=1$ when both vector norms exceed $\eps$, and $\chi=0$
otherwise. The stabilized determinant surrogate and response are
\begin{equation}
g=\max\{\eps,1-c^2\},
\qquad
\boxed{R_D(\bm u,\bm v)=\chi g^{D/4}.}
\label{eq:normalized_response}
\end{equation}
For a valid pair and ignoring regularization, $c=\cos\theta$ and
\begin{equation}
\boxed{R_D=\abs{\sin\theta}^{D/2}.}
\label{eq:angular_response}
\end{equation}
Removing the vector lengths makes this response dimensionless. The geometric
$\lambda^D$ law in Equation~\eqref{eq:magnitude_scaling} therefore does not
describe the normalized activation under image zooming or intensity scaling.
If either vector has near-zero norm, its angle is undefined. The implementation
multiplies \eqref{eq:normalized_response} by a validity mask and follows the
zero-frame convention instead of assigning a maximal response to a zero
vector.
For fixed $D>0$, $g\mapsto g^{D/4}$ is strictly increasing on $(0,1]$.
Consequently, a fixed-$D$ pointwise response does not contain more information
than the underlying angular Gram term $g$. Any practical effect must instead
arise from parameterized reweighting, spatial aggregation, or interaction with
the downstream model. Distinguishing these mechanisms requires controlled
comparisons with raw $g$, $\abs{\sin\theta}$, and a generic learnable power
activation.

\subsection{Differentiation and Near-Degenerate Gradients}

For a valid pair and fixed $g\in(0,1]$,
\begin{equation}
\boxed{\frac{\partial R_D}{\partial D}
=\frac14\log(g)g^{D/4}.}
\label{eq:gradient_d}
\end{equation}
Thus $\partial R_D/\partial D<0$ when $0<g<1$: increasing $D$ suppresses the
normalized response, whereas decreasing $D$ increases it.

Without the numerical floor, $R_D=(1-c^2)^{D/4}$ and
\begin{equation}
\boxed{\frac{\partial R_D}{\partial c}
=-\frac{D}{2}c(1-c^2)^{D/4-1}.}
\label{eq:gradient_c}
\end{equation}
As $g\to0^+$, $g^{D/4}\log g\to0$ for every $D>0$, so the exact
$D$-gradient tends to zero despite the logarithm. In contrast, the
$c$-gradient can diverge as $\abs{c}\to1$ when $0<D<4$, remain finite at
$D=4$, and vanish for $D>4$. Computing
$R_D=\exp[(D/4)\log g]$ with a documented $\eps$ floor avoids undefined
logarithms and limits the gradient. A hard clamp is nondifferentiable at its
floor and has zero derivative below it; smooth floors are an alternative when
gradient continuity is more important than exact clipping.

\subsection{Constrained Learnable Exponent}

The assumption $D>0$ is enforced through the parameterization
\begin{equation}
\boxed{D_{c,s}=D_{\min}+(D_{\max}-D_{\min})\sigma(\rho_{c,s}),}
\label{eq:constrained_d}
\end{equation}
where $c$ indexes channels, $s$ indexes patch scales, and $\sigma$ is the
logistic function. The implementation uses
$D_{\min}=0.1$, $D_{\max}=3.0$, and initializes $D_{c,s}=1.5$. Restricting
$D$ to $(1,2)$ could encode a retinal-vascular prior
\cite{Masters2004}. The broader interval is appropriate here because the
parameter is treated as a task-adaptive response exponent rather than a
physical vascular dimension.

\subsection{Multi-Scale FCPLayer}

Given an image tensor $\bm I\in\R^{C\times H\times W}$ and an even patch width
$s$, the implementation extracts an $s\times s$ neighborhood at each pixel.
Within every channel, the top and bottom half-rows are averaged to form
$\bm u,\bm v\in\R^s$, after which \eqref{eq:normalized_response} is evaluated
for each $s\in\mathcal S$. The resulting $C\abs{\mathcal S}$ maps are
concatenated and fused by a $1\times1$ convolution, batch normalization, and a
rectified linear unit. Explicit asymmetric ``same'' padding for even kernels
avoids an $(H+1)\times(W+1)$ output followed by a top-left crop.

\subsection{Non-Circular Scale Regression and Accumulation}
\label{sec:scale_regression}

The dense FCPLayer applies a prescribed or learned power independently at each
scale; it does not estimate a scaling law. A second realization therefore
estimates an exponent directly from raw responses without applying a
$D$-dependent power. For each non-overlapping patch $p$, scale $s$, channel
$c$, and split
orientation $o\in\{\mathrm{horizontal},\mathrm{vertical}\}$, paired half-patch
means define $\bm u_{p,s,c,o},\bm v_{p,s,c,o}\in\R^s$. Let
\begin{equation}
\begin{aligned}
c_{p,s,c,o}&=\operatorname{clip}\!\left(
\widehat{\bm u}_{p,s,c,o}^{\top}\widehat{\bm v}_{p,s,c,o},-1,1\right),\\
g_{p,s,c,o}&=\chi_{p,s,c,o}(1-c_{p,s,c,o}^2)
\bm 1\{1-c_{p,s,c,o}^2>\delta\},
\end{aligned}
\label{eq:raw_scale_g}
\end{equation}
where $\chi$ excludes undefined zero-norm frames and $\delta$ removes responses
at the numerical floor. Two orthogonal splits reduce directional bias. Define
the valid-frame mean and accumulated response
\begin{equation}
A_{x,c}(s)=\frac{\sum_{p,o}g_{p,s,c,o}}{\sum_{p,o}\chi_{p,s,c,o}},
\qquad
T_{x,c}(s)=\sum_{p,o}g_{p,s,c,o}.
\label{eq:scale_aggregates}
\end{equation}
For centered $\ell_s=\log s-\abs{\mathcal S}^{-1}\sum_{q\in\mathcal S}\log q$,
ordinary least squares gives
\begin{equation}
\beta_{x,c}=\frac{\sum_s\ell_s\log A_{x,c}(s)}{\sum_s\ell_s^2},
\qquad
\tau_{x,c}=\frac{\sum_s\ell_s\log T_{x,c}(s)}{\sum_s\ell_s^2}.
\label{eq:scale_slopes}
\end{equation}
Equation~\eqref{eq:scale_aggregates} defines $A$ only when at least one frame
is valid. We call $\beta_{x,c}$ the mean-response exponent and
\begin{equation}
\boxed{D_{\mathrm{acc},x,c}=-\tau_{x,c}}
\label{eq:dacc}
\end{equation}
the empirical FCP accumulation exponent. No target value of $D$ appears from
Equation~\eqref{eq:raw_scale_g} through Equation~\eqref{eq:scale_slopes}, so
the estimation is non-circular with respect to the exponent being fitted.

Its connection to self-similarity is conditional and explicit. If responding
frames satisfy $N(s)=Cs^{-D_*}$ and their conditional mean response is a
positive scale-independent constant $\mu$, then
$T(s)=\mu N(s)=\mu Cs^{-D_*}$ and \eqref{eq:dacc} recovers $D_*$. If instead
the mean response scales as $s^\gamma$, then
$D_{\mathrm{acc}}=D_*-\gamma$. If all frames are parallel, $T(s)=0$ and the
exponent is not identifiable. Thus $D_{\mathrm{acc}}$ combines changes in the
number of responding frames with changes in their conditional response
magnitude, and is not automatically a Hausdorff or box-counting dimension.
The differentiable branch retains, per channel, $\beta$, $D_{\mathrm{acc}}$,
the centered mean-response intercept,
both regression $R^2$ values, and the mean valid-frame fraction, and projects
these 18 RGB summaries to 32 features. A raw-scale control instead projects
$\log A$, $\log T$, response standard deviation, and valid fraction at every
scale. All operations, including the closed-form regressions, are
differentiable away from the documented floors. The trained branch uses
$\delta=10^{-8}$; the binary-set and untrained medical audits use $10^{-12}$.
When a scale has no valid or positive response, the code substitutes
$\log\delta$ and records zero $R^2$; a zero returned slope in that case is a
sentinel, not an estimated dimension.

\begin{figure}[t]
\centering
\definecolor{fcpnavy}{HTML}{264653}
\definecolor{fcpblue}{HTML}{457B9D}
\definecolor{fcpteal}{HTML}{2A9D8F}
\definecolor{fcpamber}{HTML}{E9A23B}
\definecolor{fcpcoral}{HTML}{E76F51}
\definecolor{fcplight}{HTML}{F5F8FA}
\resizebox{\textwidth}{!}{%
\begin{tikzpicture}[
  x=1cm,y=1cm,
  stage/.style={draw=#1,line width=0.8pt,rounded corners=3pt,
    fill=#1!10,minimum width=2.35cm,minimum height=1.02cm,
    text width=2.12cm,align=center,inner sep=3.5pt,
    font=\footnotesize,text=fcpnavy},
  compact/.style={draw=#1,line width=0.8pt,rounded corners=3pt,
    fill=#1!10,minimum width=2.55cm,minimum height=0.78cm,
    text width=2.30cm,align=center,inner sep=3pt,
    font=\footnotesize,text=fcpnavy},
  flow/.style={-{Stealth[length=5.5pt,width=5pt]},line width=0.9pt,draw=fcpnavy},
  paneltitle/.style={font=\bfseries\small,text=fcpnavy,anchor=west},
  note/.style={font=\scriptsize,text=fcpnavy,align=center}]
\filldraw[fill=fcplight,draw=fcpblue!55,line width=0.7pt,rounded corners=5pt]
  (-0.35,4.55) rectangle (17.25,6.95);
\filldraw[fill=fcpteal!3,draw=fcpteal!55,line width=0.7pt,rounded corners=5pt]
  (-0.35,2.05) rectangle (17.25,4.35);
\filldraw[fill=fcpcoral!3,draw=fcpcoral!50,line width=0.7pt,rounded corners=5pt]
  (-0.35,-0.55) rectangle (17.25,1.80);

\node[paneltitle] at (-0.12,6.67) {(a) Raw angular Gram responses without an inserted exponent};
\node[stage=fcpblue] (patch) at (0.95,5.60)
  {Non-overlapping patch\\$P_{p}^{(s)}\!\in\!\R^{s\times s}$};
\node[stage=fcpblue] (vectors) at (4.00,5.60)
  {Horizontal + vertical\\paired half means};
\node[stage=fcpteal] (normalize) at (7.05,5.60)
  {$\ell_2$ normalization\\$\widehat{\bm u},\widehat{\bm v}$};
\node[stage=fcpamber] (gram) at (10.10,5.60)
  {Angular Gram term\\$g=1-c^2$};
\node[stage=fcpcoral] (power) at (13.15,5.60)
  {Validity mask\\zero-norm frames excluded};
\node[stage=fcpteal] (response) at (16.00,5.60)
  {Raw responses\\$\{g_{p,s,c,o}\}$};
\draw[flow] (patch)--(vectors); \draw[flow] (vectors)--(normalize);
\draw[flow] (normalize)--(gram); \draw[flow] (gram)--(power);
\draw[flow] (power)--(response);
\node[note,below=1mm of power] {no $D$ appears before scale regression};

\node[paneltitle] at (-0.12,4.08) {(b) Accumulation and explicit log-log scale regression};
\node[stage=fcpblue] (featuremap) at (0.95,3.10)
  {Input tensor\\$\bm I\!\in\!\R^{C\times H\times W}$};
\node[stage=fcpteal] (scales) at (4.00,3.10)
  {Patch widths\\$s\in\{8,16,32,64\}$};
\node[stage=fcpteal] (maps) at (7.05,3.10)
  {Scale aggregates\\$A_x(s),\ T_x(s)$};
\node[stage=fcpamber] (fusion) at (10.10,3.10)
  {Closed-form OLS\\versus $\log s$};
\node[stage=fcpblue] (pool) at (13.15,3.10)
  {$D_{\rm acc}=-\tau$\\$\beta$, $R^2$, validity};
\node[stage=fcpcoral] (embedding) at (16.00,3.10)
  {Projected embedding\\$\bm z_{\rm scale}$};
\draw[flow] (featuremap)--(scales); \draw[flow] (scales)--(maps);
\draw[flow] (maps)--(fusion); \draw[flow] (fusion)--(pool);
\draw[flow] (pool)--(embedding);

\node[paneltitle] at (-0.12,1.53) {(c) Parallel medical-image classifier};
\node[compact=fcpblue] (image) at (1.00,0.48) {Medical image\\$\bm I$};
\node[compact=fcpblue] (cnn) at (4.55,0.92) {CNN backbone\\$\bm z_{\rm CNN}$};
\node[compact=fcpteal] (fcpbranch) at (4.55,0.02)
  {Scale-FCP branch\\$\bm z_{\rm scale}$};
\node[compact=fcpamber] (concat) at (8.50,0.48)
  {Feature fusion\\$[\bm z_{\rm CNN};\bm z_{\rm scale}]$};
\node[compact=fcpcoral] (classifier) at (12.20,0.48)
  {Classifier\\$\bm W\bm z+\bm b$};
\node[compact=fcpteal] (prediction) at (15.80,0.48)
  {Probabilities\\$p(y\mid\bm I)$};
\draw[flow] (image.east)--++(0.55,0)|-(cnn.west);
\draw[flow] (image.east)--++(0.55,0)|-(fcpbranch.west);
\draw[flow] (cnn.east)--++(0.55,0)-|(concat.north west);
\draw[flow] (fcpbranch.east)--++(0.55,0)-|(concat.south west);
\draw[flow] (concat)--(classifier); \draw[flow] (classifier)--(prediction);
\end{tikzpicture}
}
\caption{Scale-consistent FCP image workflow. Panel (a) computes the
dimensionless Level-III angular Gram term in two patch orientations without
inserting $D$. Panel (b) separately regresses mean and accumulated responses
over scale; $D_{\rm acc}$ is an empirical image exponent, not automatically a
Hausdorff dimension. Panel (c) fuses the projected descriptor with a CNN.}
\label{fig:fcp_pipeline}
\end{figure}
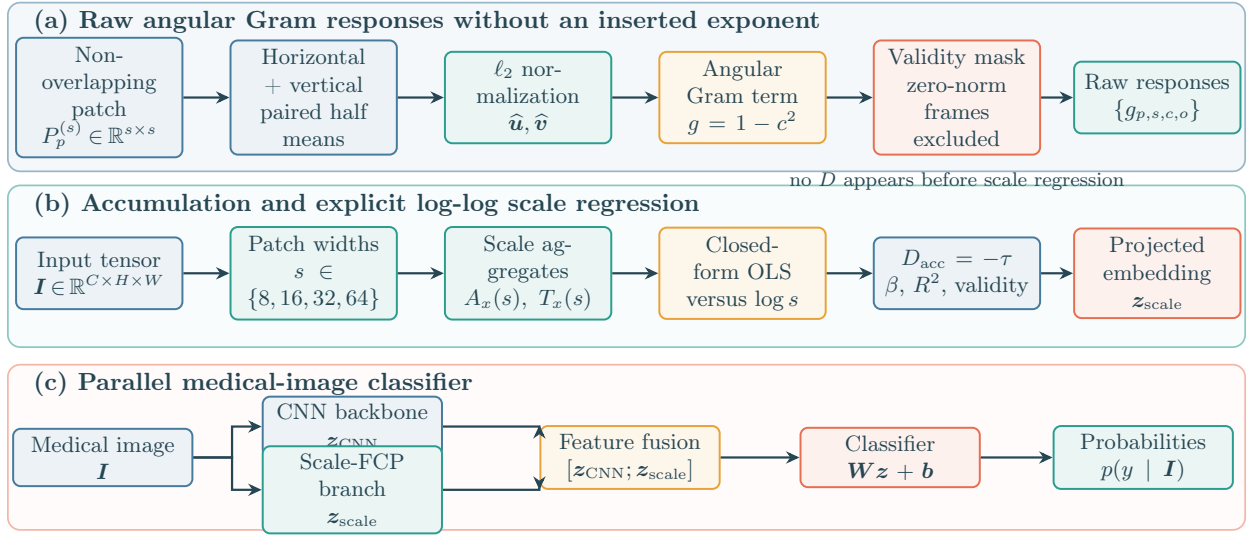

\subsection{Interpretation of Learned \texorpdfstring{$D$}{D}}

The parameters $D_{c,s}$ are shared across the dataset and are therefore
global model parameters rather than patient-specific quantities. We refer to
them as \emph{task-adaptive geometric exponents}. An independently measured
image-level box dimension $D_{\rm box}(x)$ can be tested for association with
disease severity \cite{Grauslund2010}, but such an association would not imply
$D_{c,s}=D_{\rm box}(x)$. Interpreting $D$ as a biomarker would require an
architecture that predicts an image-specific $D(x)$, calibration against an
independently validated fractal estimator, and external clinical validation.

\section{Numerical Consistency Checks}
\label{sec:numerical}

The deterministic implementation check in
\texttt{theory\_validation.py} used NumPy binary64 arithmetic, seed 20260821,
100 well-conditioned frames per case, and $k=1,\ldots,5$. Signed maximal
minors, signed log determinants, and determinant-corrected QR rotations
provided independent numerical constructions. At $D=k$, FCP and the
signed-minor product agreed exactly in the reported calculations. Across 500
frames, the maximum relative error in the Gram magnitude was
$1.25\times10^{-13}$, and the maximum normalized orthogonality residual was
$7.58\times10^{-16}$.

For $D\in\{0.75,\log3/\log2,2,2.75\}$ and 17 values of
$\lambda\in[10^{-2},10^2]$, we measured
\begin{equation}
R(\lambda)=\frac{\norm{\FCP_D(\lambda\bm v_1,\ldots,\lambda\bm v_k)}}
{\norm{\FCP_D(\bm v_1,\ldots,\bm v_k)}}.
\end{equation}
The fitted slopes differed from $D$ by at most $8.88\times10^{-16}$; the
largest relative scaling discrepancy was $3.59\times10^{-15}$. Across the
same dimensions and exponents, the maximum
rotation-equivariance residual
\begin{equation}
E_{\rm rot}=
\frac{\norm{\FCP_D(Q\bm v_1,\ldots,Q\bm v_k)-Q\FCP_D(\bm v_1,\ldots,\bm v_k)}}
{\norm{\FCP_D(\bm v_1,\ldots,\bm v_k)}}
\end{equation}
was $6.08\times10^{-15}$. Sierpi\'nski levels $q=0,\ldots,12$ deviated from
critical balance by at most $2.00\times10^{-15}$, while exponents
$D_S\pm0.35$ produced the predicted decay and growth. The angle sweep
reproduced $R_D(\theta)=\abs{\sin\theta}^{D/2}$ and the monotonicity in
\eqref{eq:gradient_d}. These calculations verify consistency between the
derived identities and their implementation; they are not independent
evidence for a fractal measure or a dimension estimator.

We then evaluated the image-level construction in
Section~\ref{sec:scale_regression} on canonical binary sets.
Table~\ref{tab:canonical_scale} reports the finest $512\times512$ raster; the
same four scales
$\{8,16,32,64\}$ were used at every resolution. The exact Sierpi\'nski
triangle has a scale-independent local angular response and a copy count that
changes by three when scale doubles, so $D_{\rm acc}$ recovered
$\log3/\log2$ to floating-point precision with $R^2=1$ at 128, 256, and 512
pixels. For the carpet, $D_{\rm acc}$ changed from 1.7640 to 1.8166 to 1.8449
as resolution increased, approaching but not reaching the similarity
dimension 1.8928. Koch-curve estimates were lower and depended on resolution.
Straight and filled sets produce parallel local frames and zero angular
response, leaving their accumulation exponent undefined even though box
counting remains possible. This non-identifiability is a structural limitation
of the response rather than a missing value to be imputed.

\begin{table}[t]
\centering
\caption{Canonical-set validation at $512\times512$. ``n.i.'' means not
identifiable because every angular response is zero.}
\label{tab:canonical_scale}
\scriptsize
\setlength{\tabcolsep}{3pt}
\begin{tabular}{@{}lcccc@{}}
\toprule
\textbf{Set} & \textbf{Known $D$} & \textbf{$D_{\rm box}$} &
\textbf{$D_{\rm acc}$} & \textbf{$R^2_{\rm acc}$}\\
\midrule
Line                 & 1.0000 & 0.9630 & n.i.   & n.i.\\
Koch curve           & 1.2619 & 1.3482 & 1.1535 & 0.9933\\
Sierpi\'nski triangle & 1.5850 & 1.5850 & 1.5850 & 1.0000\\
Sierpi\'nski carpet   & 1.8928 & 1.8765 & 1.8449 & 0.9966\\
Filled square        & 2.0000 & 1.8944 & n.i.   & n.i.\\
\bottomrule
\end{tabular}
\end{table}

The estimator is also noise-sensitive. With 20 fixed-seed replicates, flipping
only 0.25\% of Sierpi\'nski-triangle pixels changed mean $D_{\rm acc}$ from
1.5850 to $0.9271\pm0.0088$, although the fitted power law retained a high
$R^2$ ($0.9645\pm0.0076$). A high $R^2$ therefore indicates approximate
power-law behavior of the chosen response, not accuracy of the estimated
exponent.

\section{Medical Imaging Experiments}
\label{sec:experiments}

Two medical-imaging experiments examine whether the FCP construction can be
embedded in a classification pipeline. RIM-ONE retinal fundus images are used
for glaucoma classification, and FracAtlas radiographs are used for fracture
classification. In each experiment, the same compact CNN is trained from
scratch either alone or with an FCP-centered multiscale descriptor package.
The experiments assess application-level utility; the mathematical properties
of FCP do not depend on their outcomes.

\subsection{Datasets, Models, and Validation Protocol}

The original RIM-ONE DL and FracAtlas reports describe the respective image
acquisition procedures, cohorts, annotations, and class distributions
\cite{Fumero2020,Abedeen2023}.

We used the official RIM-ONE random and hospital-separated partitions and a
stratified 65/15/20 image-level split for FracAtlas, whose available metadata
do not permit a patient-disjoint guarantee. Images were resized to
$128\times128$ pixels, and preprocessing and model selection used only the
training and validation data. Each comparison used five fixed seeds and a
decision threshold of 0.5; results are reported as mean $\pm$ sample standard
deviation. Here, fusion denotes the complete descriptor package rather than
an isolated FCP-only intervention.

\section{Medical-Imaging Results}
\label{sec:results}

\subsection{FCP-Centered Fusion Performance}

The FCP-centered descriptor package achieved a higher mean AUROC than the same
CNN trained from scratch in all three five-seed comparisons. On RIM-ONE, the
differences were 0.0872 for the random partition and 0.0922 for the hospital
partition; the corresponding differences in balanced accuracy were 0.0884 and
0.0480. On FracAtlas, mean AUROC increased by 0.0072, balanced accuracy by
0.0535, and F1 by 0.0788. Table~\ref{tab:medical_fusion} also reports the
operating-point metrics, which reveal sensitivity trade-offs not captured by
AUROC. Taken together, these comparisons indicate that the complete
FCP-centered representation can complement a representation learned from
scratch on these data.

\begin{table}[t]
\centering
\caption{Five-seed FCP-centered fusion results (mean $\pm$ sample standard
deviation). Each comparison uses the same CNN trained from scratch; fusion
adds normalized FCP and complementary fractal descriptors. BAcc denotes
balanced accuracy. Bold indicates the higher mean within each two-row
comparison and is descriptive, not a statistical-significance claim.}
\label{tab:medical_fusion}
\scriptsize
\setlength{\tabcolsep}{2pt}
\resizebox{\textwidth}{!}{%
\begin{tabular}{@{}llcccccc@{}}
\toprule
\textbf{Dataset/split} & \textbf{Method} & \textbf{AUROC} &
\textbf{Accuracy} & \textbf{BAcc} & \textbf{Sensitivity} &
\textbf{Precision} & \textbf{F1}\\
\midrule
\multirow{2}{*}{FracAtlas}
& Scratch CNN
& $0.7403\pm0.0272$ & $0.6208\pm0.2431$ & $0.6186\pm0.0615$
& $\bm{0.6153\pm0.3222}$ & $0.3600\pm0.1645$ & $0.3669\pm0.0855$\\
& CNN + FCP-centered fusion
& $\bm{0.7475\pm0.0228}$ & $\bm{0.7745\pm0.0473}$
& $\bm{0.6721\pm0.0184}$ & $0.5139\pm0.1047$
& $\bm{0.4153\pm0.0777}$ & $\bm{0.4457\pm0.0187}$\\
\midrule
\multirow{2}{*}{RIM-ONE random}
& Scratch CNN
& $0.7264\pm0.0217$ & $0.6644\pm0.0366$ & $0.6526\pm0.0391$
& $0.6115\pm0.1369$ & $0.5289\pm0.0404$ & $0.5599\pm0.0633$\\
& CNN + FCP-centered fusion
& $\bm{0.8135\pm0.0097}$ & $\bm{0.7616\pm0.0276}$
& $\bm{0.7410\pm0.0295}$ & $\bm{0.6692\pm0.0516}$
& $\bm{0.6658\pm0.0436}$ & $\bm{0.6664\pm0.0373}$\\
\midrule
\multirow{2}{*}{RIM-ONE hospital}
& Scratch CNN
& $0.5843\pm0.0670$ & $0.5609\pm0.0734$ & $0.5393\pm0.0440$
& $\bm{0.4786\pm0.0636}$ & $0.3715\pm0.0534$ & $0.4134\pm0.0307$\\
& CNN + FCP-centered fusion
& $\bm{0.6765\pm0.0298}$ & $\bm{0.6299\pm0.0306}$
& $\bm{0.5873\pm0.0185}$ & $0.4679\pm0.0293$
& $\bm{0.4346\pm0.0393}$ & $\bm{0.4490\pm0.0164}$\\
\bottomrule
\end{tabular}
}
\end{table}

\subsection{RIM-ONE Retinal Application}
\label{sec:rimone_application}

On RIM-ONE, mean AUROC was higher by 0.0872 on the random partition and by
0.0922 on the hospital-separated partition. The corresponding differences in
balanced accuracy were 0.0884 and 0.0480, and those in F1 were 0.1065 and
0.0355. Every reported mean was higher for fusion on the random partition. On
the hospital-separated partition, however, mean sensitivity decreased by
0.0107 while AUROC, accuracy, balanced accuracy, precision, and F1 increased.
The hospital result is relevant to site-related generalization, but it remains
an evaluation on an official partition rather than external clinical
validation. The smaller AUROC standard deviations for fusion describe these
five runs and are not interval estimates.

\subsection{FracAtlas Fracture Application}
\label{sec:fracatlas_application}

On FracAtlas, mean AUROC was higher by 0.0072, balanced accuracy by 0.0535, and
F1 by 0.0788, while sensitivity was lower by 0.1014. Because fractures account
for 17.6\% of the data, raw accuracy alone can obscure minority-class behavior;
balanced accuracy, precision, sensitivity, and F1 must be considered together
\cite{Saito2015}. Fusion increased mean precision from 0.3600 to 0.4153 as
sensitivity decreased from 0.6153 to 0.5139. The result therefore reflects a
different operating-point trade-off rather than uniformly better fracture
detection.

Across the three comparisons, the complete FCP-centered descriptor package
gave higher mean AUROC and lower AUROC standard deviation over the five seeded
runs. The scope of attribution is narrower than the observed package-level
utility: non-FCP descriptors were included, the fusion head was larger, paired
confidence intervals could not be computed from every archived run, and
FracAtlas lacks auditable patient grouping. The experiments therefore
demonstrate the application of FCP-centered fusion without identifying the
isolated causal effect of FCP.

\section{Discussion}
\label{sec:discussion}

The mathematical contribution is a degree-controlled radial deformation of
generalized Gram volume. Within the stated Gram-power ansatz, the homogeneity
requirement determines the exponent, while the Hodge construction supplies a
canonical orientation in codimension one. Orthogonality, alternation, and
equivariance are inherited from the classical generalized cross product rather
than established as unrelated new phenomena. What changes is the homogeneity
degree and, except at $D=k$, multilinearity is lost. The self-similar balance
motivates the term \emph{fractal}, but remains an algebraic consequence of
exact frame correspondence and scaling assumptions, not a convergence theorem
for a fractal measure.

Normalization changes the object used in the image model. FCPLayer produces a
dimensionless angular power response rather than the exact length-homogeneous
vector FCP. Since every fixed positive power is a monotone bijection of $g$,
the pointwise transformation introduces no information beyond the underlying
Gram response. Any practical effect must arise through the multiscale
construction, aggregation, response conditioning, or interaction with the CNN
representation. Likewise, a learned $D$ remains a global task parameter unless
the architecture predicts it separately for each image; the present parameters
do not constitute patient-level biomarkers.

Accumulation across patch widths provides a direct connection to scale without
turning the normalized response into Hausdorff measure. The Sierpi\'nski
triangle is the favorable exact case: a stable nonzero local response and
exact copy-count scaling yield the known exponent. The carpet and Koch results
show finite-raster bias, while the line, square, and noise experiments expose
non-identifiability and instability. In particular, $D_{\rm acc}$ combines the
scaling of the responding-frame count with the scaling of conditional response
magnitude. Equation~\eqref{eq:dacc} therefore defines a testable image
descriptor and a conditional similarity-dimension estimator, not a universal
formula for fractal dimension.

The application results had a consistent direction across the two imaging
modalities and all three five-seed comparisons: the complete FCP-centered
package achieved higher mean AUROC than the identically trained CNN. The two
RIM-ONE differences were 0.0872 and 0.0922. On FracAtlas, the smaller AUROC
difference of 0.0072 accompanied differences of 0.0535 in balanced accuracy
and 0.0788 in F1, together with lower sensitivity at the fixed threshold. The
AUROC standard deviation was also lower for fusion in each comparison, though
these descriptive five-seed summaries are not confidence intervals. The
results support further study of FCP-centered fusion for representations
learned from scratch on modest medical-image cohorts.

At the same time, the application experiments evaluate a complete descriptor
package. They establish that the proposed construction can be implemented on
high-dimensional image patches, fused with CNN features, and trained
repeatedly in two medical-imaging settings. Complementary fractal descriptors
and a larger fusion input remain alternative explanations for the observed
differences. A parameter-matched comparison of the raw Gram response, FCP
alone, the complementary descriptors alone, and their combination is needed
to isolate the contribution of FCP.

\section{Limitations}
\label{sec:limitations}

The analysis does not establish equivalence with Hausdorff measure, convergence
of frame accumulations, statistical consistency of a dimension estimator, or
perturbation bounds near rank deficiency. Its exact self-similarity assumptions
exclude overlapping, anisotropic, random, and multifractal systems. At the
neural level, normalization discards length information, and a generic learned
power is a reparameterization of learned $D$. The accumulation exponent is
undefined when the angular response vanishes, becomes biased when conditional
response magnitude varies with scale, and was highly sensitive to sparse pixel
noise. Four raster scales also leave only two residual degrees of freedom,
so a high $R^2$ can coexist with a biased exponent.

The experimental evidence is also limited. RIM-ONE is small, and inputs were
reduced to $128\times128$ pixels. Complete patient identifiers were unavailable
for FracAtlas. Although the results cover five seeds, formal paired confidence
intervals and hypothesis tests could not be recovered from every archived run.
The study includes no independent external cohort, reader study, decision-curve
analysis, or clinically selected operating point. In addition, the
FCP-centered models combine several descriptor families and use a larger
fusion input, precluding attribution of the observed differences to FCP alone.
The FracAtlas split is image-disjoint but cannot be shown to be
patient-disjoint, and its sensitivity trade-off requires evaluation at
clinically selected thresholds. These experiments establish computational
applicability rather than clinical readiness; patient-grouped external
validation remains necessary.

\section{Conclusion}

We formulated a generalized Fractal Cross Product as a prescribed-degree power
of Gram volume, with a canonically oriented vector form in codimension one.
The construction recovers the generalized Euclidean cross product at $D=k$
and satisfies the stated conditional self-similar balance. Its image
realization provides a differentiable multiscale angular response, while
explicit regression of raw mean and accumulated responses yields an
image-specific empirical exponent without first inserting $D$. The canonical
experiments recovered the specified Sierpi\'nski-triangle exponent and also
revealed finite-raster bias, degeneracy, and substantial noise sensitivity.

The main application result is that FCP-centered fusion achieved higher mean
AUROC than the same from-scratch CNN in every five-seed comparison: 0.8135
versus 0.7264 on the RIM-ONE random partition, 0.6765 versus 0.5843 on the
hospital-separated partition, and 0.7475 versus 0.7403 on FracAtlas. On
FracAtlas it also
raised balanced accuracy from 0.6186 to 0.6721 and F1 from 0.3669 to 0.4457.
The repeated pattern across two imaging modalities supports further evaluation
of the complete fusion framework, but does not show that FCP alone caused the
differences or establish clinical generalization. The next steps are a
parameter-matched isolation of the FCP branch and evaluation at native
resolution on patient-grouped, multi-institutional cohorts.

\appendix

\section{Additional Generalized-FCP Analysis}
\label{app:generalized_fcp}

This appendix records supporting exterior-algebra details for the main
derivation. In an oriented Euclidean space $\R^{k+1}$, the Hodge star
maps $\bm v_1\wedge\cdots\wedge\bm v_k$ to a one-vector and
\begin{equation}
 \norm{\bm v_1\wedge\cdots\wedge\bm v_k}^{2}=\det G.
 \label{eq:app_wedge_gram}
\end{equation}
A repeated factor makes the wedge vanish, while an odd permutation reverses
its orientation. Equation~\eqref{eq:app_wedge_gram} therefore supplies the
Gram magnitude, degeneracy, orthogonality, and alternation used in
Section~\ref{sec:theory}.

The transformation law can be stated more generally. Let $V$ contain the
spanning vectors as columns, let $A\in GL(k)$ mix the input frame, and let
$Q\in O(k+1)$ transform the ambient coordinates. Multilinearity of the wedge
and the oriented Hodge-star identity give
\begin{align}
 \FCP_D(VA)&=\operatorname{sgn}(\det A)\abs{\det A}^{D/k}\FCP_D(V),
 \label{eq:app_frame_change}\\
 \FCP_D(QV)&=\det(Q)\,Q\FCP_D(V).
 \label{eq:app_orthogonal_change}
\end{align}
Thus the scalar magnitude is invariant under orthogonal transformations,
whereas the vector is rotation equivariant and changes orientation under a
reflection. Equation~\eqref{eq:app_frame_change} also recovers alternation and
the separate-argument factor $c^{D/k}$.

Finally, with $p=D/k>0$, the vector construction is the radial map
$\Phi_p(X)=\norm{X}^{p-1}X$. Because
$\norm{\Phi_p(X)}=\norm{X}^{p}\to0$, its zero assignment is continuous for
every $D>0$. Its derivative need not remain bounded at $X=0$ when $D<k$.
Continuity of the exact geometric FCP must therefore be distinguished from
gradient stability of a differentiable numerical layer near a degenerate
frame.

\section*{Author Contributions}

Samir Brahim Belhaouari: conceptualization, mathematical formulation,
supervision, and manuscript review. Noaman Khan: methodology, software,
experiments, visualization, and initial manuscript preparation. Both authors
contributed to formal analysis, interpretation, and approval of the final
manuscript.

\section*{Data Availability}

RIM-ONE DL and FracAtlas remain available from their respective providers under
their original terms \cite{Fumero2020,Abedeen2023}. Dataset images are not redistributed.
The supporting artifact retains the available run metadata and aggregation
checks needed to audit the reported descriptive analysis. Current runners save
ordered paths, labels, fit/validation indices, and predictions on every new
run; some archived Table~\ref{tab:medical_fusion} outputs predate this expanded
schema and contain aggregate metrics only.

\section*{Conflicts of Interest}

The authors declare no conflicts of interest.

\bibliographystyle{ieeetr}
\bibliography{references}

\end{document}